\documentclass[lettersize,journal]{IEEEtran}
\usepackage{amsmath,amsfonts}
\usepackage{algorithmic}
\usepackage{algorithm}
\usepackage{array}
\usepackage[caption=false,font=normalsize,labelfont=sf,textfont=sf]{subfig}
\usepackage{textcomp}
\usepackage{stfloats}
\usepackage{url}
\usepackage{verbatim}
\usepackage{graphicx}
\usepackage{cite}
\usepackage{hyperref}
\usepackage{cleveref}   

\usepackage{makecell}

\usepackage{pifont}   
\usepackage{placeins}
\usepackage{multirow}

\usepackage{amssymb}
\usepackage{amsmath} 
\usepackage{newfloat}
\usepackage{listings}

\usepackage{array}
\usepackage{tabularx}
\usepackage{booktabs}  
\usepackage{nicefrac}
\DeclareMathOperator*{\argmin}{argmin}
\usepackage{dsfont}
\usepackage{color}
\newcommand{\ee}[1]{\mathds{E}[#1]}

\newcommand{\norm}[1]{\|#1\|}

\newtheorem{remark}{\textbf{Remark}}

\begin{document}

\title{CO-PFL: Contribution-Oriented Personalized Federated Learning for Heterogeneous Networks}


\author{Ke~Xing, 
        Yanjie~Dong, 
        Xiaoyi~Fan, 
        Runhao Zeng,
        Victor C. M. Leung,
        M. Jamal Deen,
        and~Xiping~Hu%
\thanks{Ke Xing and Xiping Hu are with Shenzhen MSU-BIT University, Shenzhen, Guangdong, China, and Beijing Institute of Technology, Beijing, China.
Yanjie Dong, Xiaoyi Fan, Runhao Zeng, and Victor C. M. Leung are with Shenzhen MSU-BIT University, Shenzhen, Guangdong, China.  
M. Jamal Deen is with AI Atlas Inc., Hamilton, Canada.
Corresponding authors: Yanjie Dong (ydong@smbu.edu.cn), Xiping Hu (huxiping@smbu.edu.cn). Part of this work was presented at ACM MobiCom 2025\cite{mobicomxk}.}%
}





\maketitle

\begin{abstract}
Personalized federated learning (PFL) addresses a critical challenge of collaboratively training customized models for clients with heterogeneous and scarce local data. Conventional federated learning, which relies on a single consensus model, proves inadequate under such data heterogeneity. Its standard aggregation method of weighting client updates heuristically or by data volume, operates under an equal-contribution assumption, failing to account for the actual utility and reliability of each client's update. This often results in suboptimal personalization and aggregation bias. To overcome these limitations, we introduce Contribution-Oriented PFL (CO-PFL), a novel algorithm that dynamically estimates each client's contribution for global aggregation. CO-PFL performs a joint assessment by analyzing both gradient direction discrepancies and prediction deviations, leveraging information from gradient and data subspaces. This dual-subspace analysis provides a principled and discriminative aggregation weight for each client, emphasizing high-quality updates. Furthermore, to bolster personalization adaptability and optimization stability, CO-PFL cohesively integrates a parameter-wise personalization mechanism with mask-aware momentum optimization. Our approach effectively mitigates aggregation bias, strengthens global coordination, and enhances local performance by facilitating the construction of tailored submodels with stable updates. Extensive experiments on four benchmark datasets (CIFAR10, CIFAR10C, CINIC10, and Mini-ImageNet) confirm that CO-PFL consistently surpasses state-of-the-art methods in in personalization accuracy, robustness, scalability and convergence stability.
\end{abstract}


\section*{Nomenclature}
\label{tab:notations}
\begin{tabbing}
	\textbf{Scalars} \quad\:\:\= \textbf{Definitions} \\
        $N$ \> The number of clients in the federal system \\
        $w$  \>  Global model parameters \\
        $w_n$ \>  Client $n$'s local model\\
        $g_n$ \> Shared submodel of client $n$ \\
        $p_n$ \>  Personalized submodel of client $n$ \\
        $\mathcal{M}_n$ \> Dataset on the client $n$ \\
        $m_n^k$ \> Parameter mask matrix of client $n$\\
        $\eta$ \> Learning rate \\
        $K$ \> Number of global training rounds \\
        $T$ \> Number of local iterations\\
        $\beta_1, \beta_2$ \> Momentum optimizer constant parameter \\
        $\gamma$ \>  Parameter personalization limit \\
        $p$ \> Parameter personalization ratio\\
        $\delta_n^k$  \> Update direction of client $n$ in round $k$ \\
        $\bar \delta_{-n}^k$ \> Client average update direction except $n$ \\
        $w_{-n}^k$  \>  Average model parameters excluding client $n$ \\
        $\zeta_n$   \>  Data samples drawn from client $n$ \\
        $\Gamma_n^{\text{grad}}$ \> Gradient-based contribution score of client $n$ \\
        $ \Gamma_n^{\text{data}}$ \> Prediction-based contribution score of client $n$ \\
        $\alpha_n^k$ \> The aggregated weight of client $n$ in round $k$ 
\end{tabbing}

\section{Introduction}

Federated learning (FL) has emerged as a promising paradigm for collaborative machine learning across decentralized and sensitive data sources. Unlike conventional centralized training schemes that require direct aggregation of raw data into a single repository, FL enables model parameters to be exchanged and optimized while data remain securely stored at the local clients \cite{mcmahan2017communication}. This distributed paradigm alleviates privacy risks and legal concerns associated with data sharing, making it particularly attractive in contexts where strict confidentiality must be maintained and continuous data collection is impractical \cite{wang2024comprehensive, 11077470}.

The practical potential of FL is vast. In mobile personalization, it allows for training intelligent services directly on users' devices such as smartphones and wearables, to deliver personalized recommendations and adaptive experiences without leaking private behavioral data to the cloud or central servers~\cite{LEE20071194, 10.1145/2382196.2382266, Xing2025}. This approach not only enhances privacy, but also reduces communication overhead, a crucial benefit for resource-constrained devices or mobile platforms.Similarly, in healthcare, FL enables collaborative model development across hospitals and research institutions, overcoming barriers arising from data sensitivity and regulatory frameworks, including the Health Insurance Portability and Accountability Act (HIPAA)~\cite{beyan2020distributed, haripriya2025privacy, Shangguan2025}. Furthermore, in the realm of edge intelligence, FL empowers heterogeneous devices at the network periphery, from smart home sensors to industrial IoT systems, to learn collectively, enabling low-latency decision-making and robust distributed intelligence~\cite{5978711, majumder2017smart, zhang2025decentralized}. These diverse applications collectively highlight the importance of FL's role as a cornerstone for building scalable, privacy-preserving AI services.

However, a fundamental challenge threatens this promise: data heterogeneity across clients. Standard, or "vanilla," FL algorithms aim to produce a single, consensus model under the assumption that the client’s data is independently and identically distributed. In reality, local datasets are often scarce and exhibit significant statistical heterogeneity, meaning their feature and label distributions vary dramatically due to divergent user behaviors, geographical locations, or data acquisition processes~\cite{9599369}. This heterogeneity severely degrades the performance of a one-size-fits-all model, leading to poor generalization and convergence issues on individual clients' unique data~\cite{ye2023heterogeneous,pei2024review}.

To tackle this, the field has pivoted towards Personalized Federated Learning (PFL), which seeks to collaboratively train models that assimilate valuable global knowledge while remaining finely attuned to local data distributions~\cite{pmlr-v162-zhang22o, 11077470}. A predominant PFL strategy involves architecturally decomposing the model into shared and personalized components~\cite{sun2021partialfed, chen2023sharper}. Early methods employed a fixed partitioning, designating certain layers (e.g., early feature extractors) as shared and others (e.g., final classifiers) as locally adaptable ~\cite{arivazhagan2019federated, collins2021exploiting,ma2022layer, oh2022fedbabu}.  While a step forward, this rigid decomposition often fails to accommodate the diverse and evolving needs of clients, especially when faced with significant distributional shifts.

\begin{figure}[t]
\centering
\includegraphics[width=0.90\columnwidth]{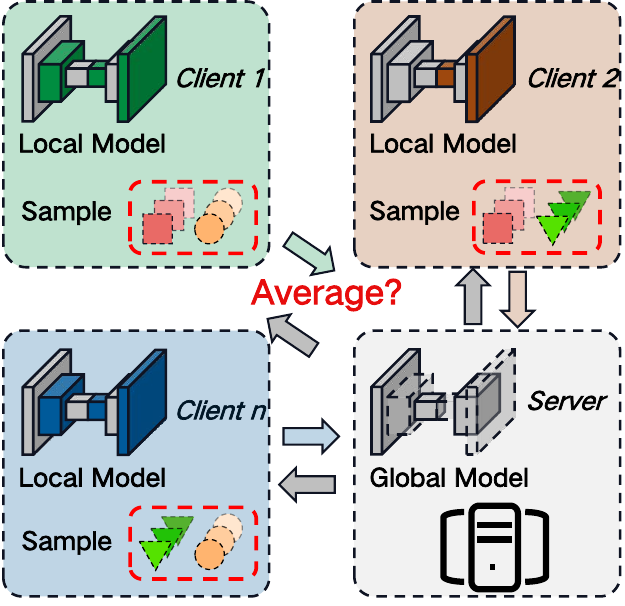} 
\caption{Conventional federated learning averages client updates uniformly, disregarding differences in data quality and model contribution, which motivates the development of contribution-weighted aggregation strategies.}    
\label{fig:f1}
\end{figure}

Seeking greater flexibility, recent research has explored dynamic sub-model selection and personalized masking mechanisms. These approaches allow clients to adaptively determine which parts of a global model to share or personalize, creating a more customized architecture for each participant~\cite{tamirisa2024fedselect, xupersonalized, tang2023learning, mclaughlin2024personalized}. As illustrated in Figure \ref{fig:f1}, these methods represent an advance. However, they share a critical, unaddressed weakness: their aggregation strategy. Despite creating unique local models, these methods typically fall back on conventional federated averaging, with uniform or data-volume-based weighting, when aggregating updates. This implicitly operates under an "equal-contribution assumption," treating all client updates as equally valuable. In a heterogeneous environment, this is a flawed premise; naïvely averaging updates from clients with varying data quality and informativeness can dilute valuable knowledge and introduce aggregation bias, ultimately hindering global collaboration. This critical shortfall motivates our investigation into a more intelligent, contribution-oriented PFL algorithm. A comparative summary of our method against related works is provided in Table \ref{tab:chayi}.

\begin{table}[t]
\centering
\caption{Comparison of recent personalized federated learning algorithms.}
\label{tab:chayi}
\renewcommand{\arraystretch}{1.25}   
\setlength{\tabcolsep}{4.5pt}        
\begin{tabular}{lccc}
\toprule
\textbf{Algorithm} &
\textbf{\makecell{Parameter \\ Personalization}} &
\textbf{\makecell{Optimizer \\  Adaptation}} &
\textbf{\makecell{Personalized \\Aggregation}} \\ 
\midrule
FedAvg~\cite{mcmahan2017communication}   & $\times$ & $\times$ & $\times$ \\
FedAvg+FT                                & $\times$ & $\times$ & $\times$ \\
LG-FedAvg~\cite{liang2020think}          & \checkmark & $\times$ & $\times$ \\
FedPer~\cite{arivazhagan2019federated}   & \checkmark & \checkmark & $\times$ \\
FedPAC~\cite{xupersonalized}             & \checkmark &  \checkmark  & $\times$ \\
FedSelect~\cite{tamirisa2024fedselect}   & \checkmark & \checkmark & $\times$ \\
\midrule
\textbf{CO-PFL}                   & \textbf{\checkmark} & \textbf{\checkmark} & \textbf{\checkmark} \\
\bottomrule
\end{tabular}
\end{table}

To overcome the limitation of equal-contribution aggregation, we propose the \textbf{C}ontribution-\textbf{O}riented \textbf{P}ersonalized \textbf{F}ederated \textbf{L}earning (CO-PFL) framework. Its core innovation is the \textbf{C}ontribution-\textbf{O}riented \textbf{W}eighted \textbf{A}ggregation (COWA) module. COWA moves beyond simple data-size-based averaging by jointly assessing two complementary metrics: gradient direction discrepancy and prediction deviation. The gradient discrepancy measures the alignment of a client's update with the overall optimization direction, helping to identify and downweight conflicting or noisy signals. Simultaneously, the prediction deviation assesses how a client's model performs on a held-out validation set, serving as a proxy for its generalization capability and informativeness. By leveraging this dual-subspace analysis, COWA assigns a discriminative aggregation weight to each client, effectively emphasizing high-quality, constructive updates while suppressing those that are biased or uninformative. This contribution-aware approach directly mitigates the adverse effects of statistical heterogeneity, leading to a more robust global model and superior personalization.

Furthermore, to ensure stable and efficient convergence within this dynamic, partial-model-sharing environment, CO-PFL integrates two supporting modules. The first is the \textbf{P}arameter-\textbf{W}ise \textbf{P}ersonalization \textbf{M}echanism (PWPM). Unlike rigid layer-level partitioning, PWPM enables each client to dynamically identify and personalize the most critical parameters based on local gradient sensitivity, all within a defined personalization budget. This allows the scope of personalization to fluidly adapt to the unique characteristics of each client's data, providing unparalleled flexibility. The second module is \textbf{M}ask-\textbf{A}ware \textbf{M}omentum \textbf{O}ptimization (MAMO). A key technical challenge in mask-based PFL is that the set of parameters being updated changes as masks evolve. MAMO stabilizes training by maintaining separate momentum buffers for shared and personalized parameters. This prevents "momentum leakage," where historical velocity from a previously personalized parameter incorrectly influences a now-shared parameter, thereby accelerating convergence and enhancing training stability in highly heterogeneous settings.

Motivated by the critical gaps in uniform aggregation and inconsistent personalization under heterogeneous federated settings, the four main contributions of this work are summarized below:

\begin{itemize}
    \item First, we propose a novel Contribution-Oriented Personalized Federated Learning (CO-PFL) framework that fundamentally rethinks the aggregation process from a contribution-aware perspective. CO-PFL is designed to achieve a principled balance between global collaboration and local adaptation by explicitly evaluating the quality of each client's update.
    
    \item Second, the core theoretical innovation is the COWA module, which quantifies client contributions through a joint analysis of gradient discrepancy and prediction deviation. This adaptive weighting scheme directly mitigates the equal-contribution bias of traditional PFL, enabling more informative and stable aggregation under heterogeneous data distributions.

    \item Third, to complement the intelligent aggregation process, we design two supporting components: (i) the PWPM, which dynamically identifies client-specific sensitive parameters based on gradient variation for fine-grained personalization, and (ii) the MAMO, which stabilizes training by decoupling momentum updates between shared and personalized submodels.

    \item Fourth, we conduct extensive experiments on four challenging benchmark datasets: CIFAR10, CIFAR10C, CINIC10, and Mini-ImageNet, which demonstrate that CO-PFL consistently surpasses state-of-the-art personalized FL algorithms in terms of personalization accuracy, robustness to distribution shifts, and convergence stability.
\end{itemize}

\section{Related Works}
The FL is a decentralized machine learning paradigm where clients perform local training on their private datasets and periodically transmit model updates to a central server for aggregation~\cite{mcmahan2017communication}. Unlike conventional paradigms such as transfer learning or multitask learning, FL explicitly enforces privacy preservation and communication efficiency by avoiding raw data sharing. However, client datasets are often heterogeneous in real-world scenarios, and such statistical heterogeneity can significantly undermine the performance of the shared model~\cite{hsieh2020non, zhang2021edge, aono2017privacy}. To mitigate these challenges, recent research attention has shifted towards PFL algorithms, which aim to tailor models to each client’s local data distribution while still leveraging the advantages of global knowledge sharing. PFL is particularly valuable in applications such as healthcare, recommendation systems, and multilingual modeling, where user-specific or institution-specific data distributions differ considerably.

\subsection{Partition-based Personalization Methods}
Early strategies to address data heterogeneity primarily relied on static architectural partitioning, where a model is manually divided into a globally shared component and a locally personalized component.

Seminal works in this category, such as FedPer~\cite{arivazhagan2019federated}  and FedRep~\cite{collins2021exploiting}, operate on a fixed decomposition. FedPer enhances the standard FedAvg framework by collaboratively training shared base layers (e.g., a feature extractor) while keeping later layers (e.g., a classifier) local to each client. While this alleviates some effects of heterogeneity, its rigid structure may not adapt to the diverse needs of all clients. FedRep follows a similar principle, personalizing only the classification head to reduce communication costs, but it assumes all clients benefit equally from a single, global feature representation. SplitNN~\cite{vepakomma2018split} employs a different kind of partition by splitting the neural network between clients and a server; clients send intermediate features to the server for the remainder of the computation. However, it does not inherently personalize models and relies on simple averaging for aggregation.

Other approaches incorporate regularization to balance local and global models. For instance, PMFL~\cite{pmlr} maintains local models that are constrained not to deviate too far from a global consensus, allowing for limited adaptation while ensuring consistency. Similarly, LG-FedAvg~\cite{liang2020think} explicitly decomposes the model into personalized and shared parts, enabling clients to learn local representations. FedBABU~\cite{oh2022fedbabu} adopts an alternative strategy by only training a shared backbone during the FL process and freezing the classification heads, which are later fine-tuned for evaluation. KT-pFL~\cite{NEURIPS2021_5383c731} emphasizes the importance of server  aggregation in PFL and it introduces a customized knowledge transfer aggregation algorithm. However, KT-pFL lacks personalized decoupling of parameters. While these methods provide foundational personalization, their aggregation strategy remains a critical weakness: the server typically employs a naive averaging rule, such as weighting by data volume, without considering the actual utility or contribution of each client's update~\cite{arivazhagan2019federated,oh2022fedbabu,liang2020think}. This overlooks the varying quality and informativeness of updates in a heterogeneous environment.

\subsection{Dynamic-based Personalization Methods}

Recognizing the limitations of fixed architectures, more recent research has explored dynamic, parameter-wise personalization, allowing clients to adaptively select which parts of a model to personalize.

For example, FedSelect~\cite{tamirisa2024fedselect} enables each client to dynamically select a personalized submodel by evaluating the importance of the parameter based on gradient. 
FedPAC~\cite{xupersonalized} aligns intermediate representations across clients while personalizing the classification heads. 
Reads~\cite{10933559} proposes a layer aggregation mechanism, however, it relies heavily on clustering outcomes and lacks granularity in its design. 
MoM-PFL~\cite{NEURIPS2024_a7a6465b} formulates each client model as a weighted mixture of multiple expert networks shared across clients. This design effectively improves adaptability across clients with diverse data patterns. However, MoM-PFL primarily focuses on mixture based prediction and model fine-tuning rather than explicitly quantifying client contributions or addressing aggregation fairness.
A parallel line of inquiry has begun to question the aggregation rule. FedALA~\cite{zhang2023fedala} adapts aggregation weights based on network architecture, but fails to leverage the contribution information embedded in the clients' local data. FedCALM~\cite{Zheng_2025_CVPR} first identified the phenomenon of deep model degradation. However, FedCALM primarily focuses on directional consistency of updates and assumes homogeneous contribution across clients, lacking an explicit estimation of client informativeness or reliability during aggregation. 
Despite this advance, it still assumes homogeneous client contributions, lacking an explicit mechanism to estimate the informativeness or reliability of each update.

Another significant branch of PFL research focuses on optimization-based personalization. 
For example, Ditto~\cite{DBLP} introduces a regularization-based approach that balances local adaptation with global consistency. 
While helping mitigate conflicts between local objectives and global convergence, Ditto still overlooks the client contributions during aggregation. 
pFedMe~\cite{NEURIPS2020_f4f1f13c} further develops this concept using a bilevel optimization formulation based on the Moreau envelope, which decouples personalization from global model learning. While these methods are highly effective for regularization-driven personalization, they personalize the entire parameter vector rather than selecting subsets, and their server-side aggregation remains a uniform average. Consequently, they do not address the core challenge of contribution-aware aggregation or provide fine-grained, parameter-wise adaptation.

In summary, while existing methods have made significant strides in personalization through architectural and optimization-based techniques, a critical gap remains. As illustrated in Table \ref{tab:chayi}, most state-of-the-art algorithms, whether partition-based or dynamic, rely on aggregation rules that fail to quantitatively assess and leverage the varying value of each client's update. This universal shortcoming, the "equal-contribution assumption," undermines global collaboration in the presence of high data heterogeneity. It is this identified gap that directly motivates our proposed CO-PFL framework, which introduces a principled, contribution-oriented aggregation module (COWA) complemented by dynamic parameter-wise personalization (PWPM) and stable optimization (MAMO) to effectively address the challenges of heterogeneous and scarce client data.

\begin{figure*}[h]
	\centering
	\includegraphics[width=1.0\textwidth]{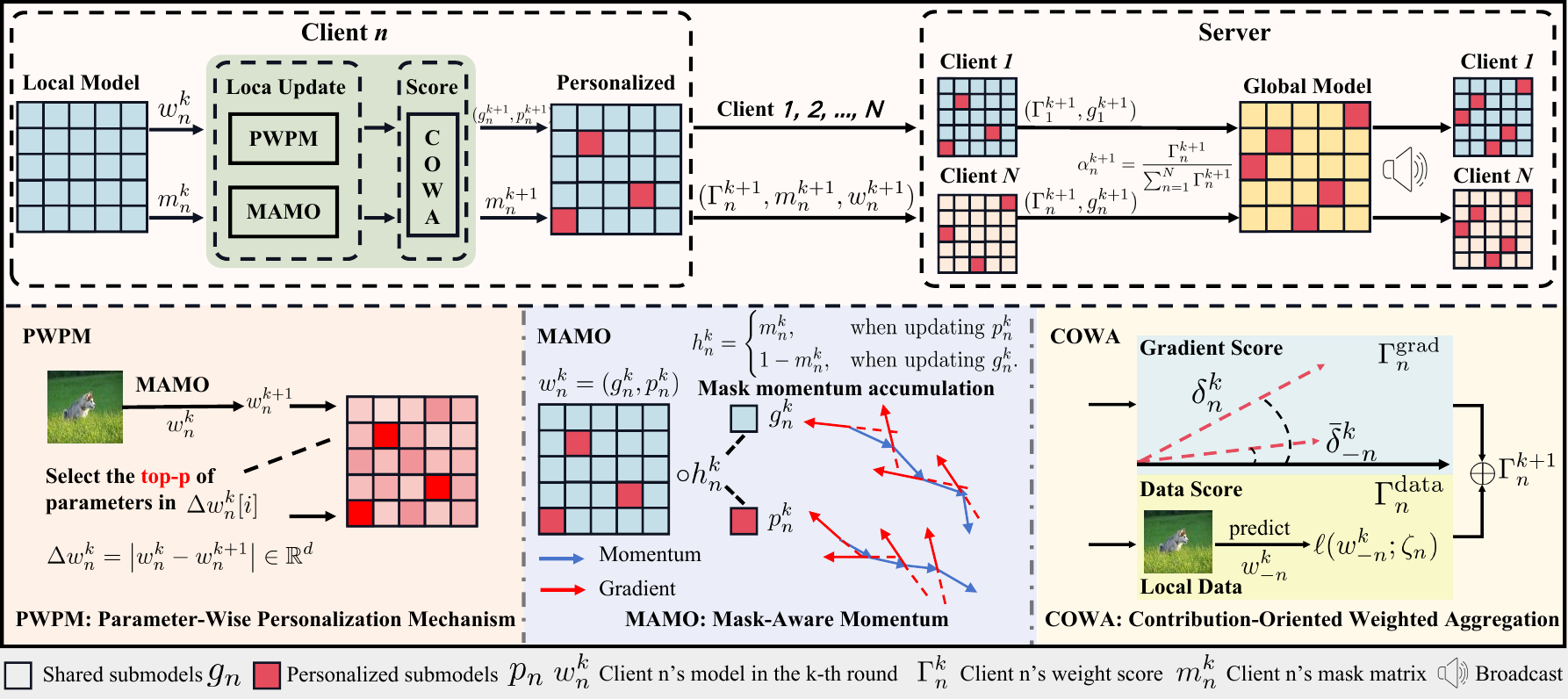}
	\caption{Overall architecture of the proposed CO-PFL framework, illustrating the client–server collaboration process and the interaction among the PWPM, MAMO, and COWA modules.}
	\label{fig:framework}
\end{figure*}

\section{Preliminaries}

In this section, we will briefly introduce the definition of PFL. Nomenclature presents the frequently used notations and corresponding descriptions throughout this paper.

\subsection{Federated Learning}
Vanilla FL aims at learning a consensus model $w \in \mathbb{R}^d$ by minimizing the population loss over all $N$ clients under the coordination of a central server. 
Denoting the local dataset per client $n$ as $\mathcal{M}_n$, the objective of vanilla FL is 
\begin{equation}\label{eq:01}
\min_{w} \frac{1}{N} \sum_{n=1}^{N}  f_n(w)
\end{equation}
where local loss function $f_n(w)$ is defined as $f_n(w) := \ee{ \ell(w; \zeta_n) }$ with $\ell(w; \zeta_n)$ as the loss incurred by model parameter $w$ on a sample $\zeta_n \sim \mathcal{M}_n$.

\begin{remark}
As shown in \eqref{eq:01}, vanilla FL produces a single consensus model based on the local data of participating clients.
\end{remark}

\subsection{Personalized Federated Learning}
However, the data distributions across clients are typically heterogeneous due to discrepancies in the acquisition processes of feature and label subspaces~\cite{10.1145/3625558}. 
Statistical heterogeneity undermines the effectiveness of the consensus model, which may fail to generalize well to the local data of individual clients~\cite{Mendieta_2022_CVPR}. 
The adverse impact becomes particularly pronounced when local datasets contain only a limited number of samples.

In order to address data heterogeneity in federated learning, different strategies have been explored to realize this objective, including: (1) regularization-based methods that constrain local models not to deviate too far from the global model~\cite{pillutla2022federated}, (2) meta-learning frameworks that learn globally transferable initializations~\cite{fallah2020personalizedfederatedlearningmetalearning}, and (3) partition-based approaches that explicitly separate shared and personalized submodels~\cite{collins2021exploiting, oh2022fedbabu}.

Within these research directions, partition-based methods are perhaps the most widely adopted \cite{pillutla2022federated, collins2021exploiting, oh2022fedbabu}. PFL extends the FL paradigm by allowing each client to maintain a personalized submodel while simultaneously learning a shared component that captures global knowledge. Formally, the objective of PFL is

\begin{equation}\label{eq:02}
\begin{split}
[w_n^*]_{n=1}^N = \argmin_{ [w_n]_{n=1}^N } &\; \frac{1}{N} \sum_{n=1}^{N} f_n( w_n  )\\
\mbox{s.t.}&\; g_1 = \ldots = g_n = \ldots = g_N
\end{split}
\end{equation}
where the local model per client $n$ can be denoted as $w_n = g_n + p_n$, $n = 1, \ldots, N$.

For instance, the classification head is trained locally on each client while the feature extractor is trained jointly in all clients \cite{collins2021exploiting, oh2022fedbabu}. 
Although such static partition strategies are simple to implement and may occasionally deliver satisfactory performance, they fail to account for the dynamic and client-specific importance of model parameters throughout the training process. 
Consequently, their adaptability becomes limited under highly heterogeneous data distributions. To overcome this limitation, we propose CO-PFL, a data-driven personalized federated learning framework that adaptively identifies the key parameters of each client during local training.
Specifically, the CO-PFL periodically determines the personalized and shared submodels by leveraging parameter contributions from both the gradient and data subspaces.

\section{Algorithmic Development of CO-PFL}
Inspired by submodel discovery (e.g., the Lottery Ticket Hypothesis~\cite{Lottery}) and recent advances in dynamic personalization~\cite{tamirisa2024fedselect}, \textbf{CO-PFL} moves beyond static pruning strategies to enable more effective and efficient learning in heterogeneous and data-scarce settings.

\subsection{Overall Description of CO-PFL}
The CO-PFL algorithm consists of three key modules: MAMO, PWPM, and COWA. 
Specifically, the MAMO module accelerates and stabilizes convergence; 
the PWPM module dynamically identifies client-specific important parameters; 
and the COWA module adjusts aggregation weights based on each client's contribution.

\begin{algorithm}[h]
	\caption{CO-PFL Algorithm}\label{alg:main}
	\begin{algorithmic}[1]
		\REQUIRE Server model $w_0$, server mask $m_0$, client masks $[m_n^0]_{n=1}^N$, training rounds $K$
		\FOR{$k = 0, 1, \ldots, K-1$}
		\STATE Server broadcasts server model $w_0^k$ and server mask $m_0^k$ to all clients \label{line2}
		\STATE Each client $n$ obtains the shared submodel, client mask, and contribution score as \label{line3}
		\begin{equation*}
			 w_n^{k+1}, m_n^{k+1}, \Gamma_n^{k+1} \leftarrow \textbf{Client}(w_0^k, m_0^k, \mathcal{M}_n)
		\end{equation*}
		\STATE Server respectively updates its mask and normalized score per client $n$ as \label{line4}
		\begin{align*}
		 &\textbf{Server Mask: }m_0^{k+1} = \bigvee_{n=1}^{N} m_n^{k+1} \\
		 &\textbf{Normalized Score: }\alpha_n^{k+1} = \frac{\Gamma_n^{k+1}}{\sum_{i=1}^N \Gamma_i^{k+1}}
		\end{align*}
		\STATE Server aggregates the shared submodel as \label{line5}
		\begin{equation*}
		g^{k+1} = \sum_{n=1}^N \alpha_n^{k+1}  w_n^{k+1} \circ (1 - m_n^{k+1})
		\end{equation*}
		\STATE Server updates its model as \label{line6}
		\begin{equation*}
			w^{k+1} = (g^{k+1} \circ (1 - m_0^{k+1}), w^k \circ m_0^{k+1})
		\end{equation*}
		\ENDFOR
		\STATE \textbf{return} Personalized models $\{w_n^K = (g_n^K, p_n^K)\}_{n=1}^N$
	\end{algorithmic}
\end{algorithm}

{\color{black} Figure~\ref{fig:framework} illustrates the overall workflow of the proposed CO-PFL algorithm. Each client $n$ maintains a local model $w_n^k$ and corresponding mask $m_n^k$, which divide the parameters into shared $g_n$ and personalized $p_n$ submodels. During local training, the PWPM dynamically selects the top-p most sensitive parameters for personalization based on local gradient variations $\Delta w_n^k$, enabling adaptive and fine-grained model customization. The MAMO module accumulates independent momentum for the shared and personalized submodels, preventing momentum leakage across masks and stabilizing optimization under evolving parameter partitions. After local updates, each client computes its COWA score $\Gamma_n^k$, which combines gradient-based contribution $\Gamma_n^{\text{grad}}$ and data-based contribution $\Gamma_n^{\text{data}}$ to quantify the reliability and informativeness of local updates. The server aggregates the shared submodels $g_n^{k+1}$ using these contribution weights to form the global model, which is then broadcast back to all clients. CO-PFL alternates between client-side local training and server-side aggregation to enable adaptive personalization and contribution-oriented collaboration. The detailed procedure is summarized in Algorithm~\ref{alg:main}.

\noindent \textit{1) \textit{Client-side Operations:}} At Step~\ref{line2} of Algorithm~\ref{alg:main}, the server broadcasts the current global model \( w_0^k \) along with its corresponding binary mask \( m_0^k \) to all participating clients. Each element \( m_0^k[i] \) is set to 1 if the associated parameter is designated for personalization.  
At Step~\ref{line3}, client \( n \) receives \( (w_0^k, m_0^k) \) and performs local training using its private dataset \( \mathcal{M}_n \). During the local optimization process, the \textit{Mask-Aware Momentum Optimization (MAMO)} module restricts momentum propagation to the personalized submodel, thereby mitigating gradient interference between the shared and personalized parameter spaces.  
Upon completing local training, the \textit{Parameter-Wise Personalization Mechanism (PWPM)} is invoked to identify the top-\(p\) parameters exhibiting the largest magnitude of local changes, which are considered the most sensitive to client-specific data. PWPM then generates a personalization mask \( m_n^k \), decomposing the updated model \( w_n^k \) into a personalized component \( p_n^k = w_n^k \circ m_n^k \) and a shared component \( g_n^k = w_n^k \circ (1 - m_n^k) \).  
Each client subsequently evaluates its local update through two metrics: the gradient-based contribution score \( \Gamma_n^{\text{grad}} \) and the prediction-based contribution score \( \Gamma_n^{\text{data}} \). These two quantities are combined into a unified contribution estimate \( \Gamma_n^{k+1} \), which, together with the updated model \( w_n^{k+1} \) and the mask \( m_n^{k+1} \), is transmitted back to the server for aggregation.

\noindent \textit{2) \textit{Server-side Operations:}} In Step~\ref{line4} of Algorithm~\ref{alg:main}, the server collects the model updates, masks, and contribution scores from all clients, denoted as \( \{w_n^{k+1}, m_n^{k+1}, \Gamma_n^{k+1}\}_{n=1}^N \). Based on the received contributions, normalized aggregation weights \( \alpha_n^{k+1} \) are computed. The global shared mask is updated through a logical OR operation across all client masks, ensuring the union of active personalized parameters.  
At Step~\ref{line5}, the server performs \textit{Contribution-Oriented Weighted Aggregation (COWA)} over the shared submodels to generate a contribution-weighted global update. Finally, in Step~\ref{line6}, the global shared submodel is refined by integrating the aggregated shared parameters with the aggregated personalized components, thereby completing one full communication round of the CO-PFL framework.

By jointly leveraging COWA, PWPM, and MAMO, the proposed CO-PFL achieves fine-grained personalization and effective collaboration across heterogeneous clients.

\subsection{Mask-Aware Momentum Module}
The optimization landscape of the PFL becomes increasingly irregular due to the existence of personalized and shared submodels. 
Applying standard optimizers uniformly over all submodels may disrupt the integrity of the parameter space when personalized and shared submodels exhibit distinct update dynamics.

\begin{algorithm}[ht]
\caption{{\bfseries Client}$(w_0^k, m_0^k, \mathcal{M}_n$)}\label{alg:local}
\begin{algorithmic}[1]
\REQUIRE Learning rate $\eta$, momentum factors $\beta_1$ and $\beta_2$, number of local iterations $T$, personalization budget $\gamma$, and personalized rate $p$
\STATE \textbf{/* MAMO-Personalized Submodel Update */}
\STATE Set $w_n^{k,0} = w_n^k$ \label{mamo:start}
\FOR{$t = 0, \ldots, T-1$}
\STATE Each client $n$ selects the personalized mask $h_n^k$ via \eqref{eq:aux} and updates the personalized submodel via \eqref{eq:moment} and \eqref{eq:update}
\ENDFOR
\STATE Each client $n$ sets $p_n^{k, T} = w_n^{k, T} \circ m_n^k$
\STATE \textbf{/* MAMO-Shared Submodel Update */}
\STATE Set $w_n^{k,0} = w_n^k$
\FOR{$t = 0, \ldots, T-1$}
\STATE Each client $n$ selects the shared mask $h_n^k$ via \eqref{eq:aux} and updates the shared submodel via \eqref{eq:moment} and \eqref{eq:update}
\ENDFOR
\STATE Each client $n$ sets $g_n^{k, T} = w_n^{k, T} \circ ( 1 -  m_n^k )$
\STATE Each client $n$ calculates $w_n^{k+1} = g_n^{k, T} + p_n^{k, T}$  \label{mamo:end}
\STATE \textbf{/* PWPM */}
\STATE Each client $n$ computes the difference as \eqref{eq:param-diff}
\STATE Each client $n$ selects top-$p$ elements of $\Delta w_n^k$ and update $m_n^{k+1}$ subject to the constraint $\norm{m_n^{k+1}} \leq \gamma d$
\STATE Each client $n$ updates the personalized and shared submodels as
\begin{equation*}
p_n^{k+1} = w_n^{k+1} \circ m_n^{k+1} \mbox{ and } g_n^{k+1} = w_n^{k+1} \circ (1 - m_n^{k+1})
\end{equation*}
\STATE \textbf{/* COWA */}
\STATE Each client $n$ updates gradient score $\Gamma_n^{\text{grad}}$ via  \eqref{gradient2}
\STATE Each client $n$ updates prediction score $\Gamma_n^{\text{pred}}$ via \eqref{space2}
\STATE Each client $n$ updates the contribution score as 
\begin{equation*}
\Gamma_n^{k+1} = \Gamma_n^{\text{grad}} + \Gamma_n^{\text{pred}}
\end{equation*}
\STATE \textbf{return} $w_n^{k+1}, m_n^{k+1}, \Gamma_n^{k+1}$
\end{algorithmic}
\end{algorithm}

At each local training step, the personalized and shared submodels are alternatively updated to handle their optimization landscape separately.
More specifically, we define an auxiliary binary mask sequence $h_n^k \in \mathbb{R}^d$ as 
\begin{equation}\label{eq:aux}
h_n^k = \begin{cases}
	m_n^k, & \mbox{when updating } p_n^k \\
	1 - m_n^k, & \mbox{when updating } g_n^k. 
\end{cases}
\end{equation}

	Our proposed MAMO module separately records the first- and second-order moment estimates for the personalized and shared submodels as 
	\begin{subequations}\label{eq:moment}
		\begin{align}
			\!\!\!\!u_n^{k,t+1} &\!=\! \beta_1 u_n^{k,t} \!+\! (1 \!-\!\beta_1) h_n^{k,t} \circ q_{n}^{k,t} \\
			\!\!\!\!v_n^{k,t+1} &\!=\! \beta_1 v_n^{k,t} \!+\! (1 \!-\! \beta_1) h_n^{k,t} \circ q_{n}^{k,t} \circ q_{n}^{k,t}
		\end{align}
	\end{subequations}
	where $u_n^{k,t}, v_n^{k,t} \in \mathbb{R}^d$, $q_{n}^{k,t} = \nabla\ell(w_n^{k,t}; \zeta_n^{k,t})$, and $\beta_1, \beta_2 \in (0, 1)$ are the momentum factors.

Based on the momentum updates \eqref{eq:moment}, we can compute the bias-corrected momentums and apply the corrected momentums to the masked submodel (e.g., shared or personalized submodels) as 
\begin{subequations}\label{eq:update}
\begin{align}
\hat u_n^{k,t} = \frac{u_n^{k,t}}{ 1-\beta_1^{kT + t} }, \hat v_n^k = \frac{v_n^{k,t}}{ 1-\beta_2^{kT + t} } \\
w_n^{k, t+1} = w_n^{k,t} - \eta h_n^{k,t} \circ \frac{ \hat u_n^{k,t} }{\sqrt{ \hat v_n^{k,t} } + \epsilon}
\end{align}
\end{subequations}
where $\epsilon$ is a small positive constant.

Based on \eqref{eq:aux}--\eqref{eq:update}, the binary mask sequence $h_n^k$ can decouple the update of shared submodel from that of the personalized submodel. 
Therefore, the personalized and shared submodels follow independent optimization trajectories without mutual interference.
Moreover, the recursions in \eqref{eq:aux}--\eqref{eq:update} also enable an alternating update between the personalized and shared submodels that can stabilize convergence under the heterogeneous and scarce local data.

\subsection{Parameter-Wise Personalization Module}
As shown in Algorithm \ref{alg:local}, a crucial step toward effective personalization is identifying submodel parameters that are sensitive to the local distribution.
Different from the heuristic static partition of personalized and shared submodels, we develop a PWPM module that adaptively selects submodels based on local training behavior. 
More specifically, the personalized mask $m_n^k$ evolves over training rounds based on the magnitude of each parameter update in order to enable fine-grained adaptation to client-specific needs.

The personalized mask is initialized as zero, i.e., $m_n^0 = m_0^0 = 0$ to secure that all model parameters can be globally updated at the beginning of training. 
Per each round $k$, each client $n$ alternatively performs the updates of personalized and shared submodels  as shown in lines \ref{mamo:start}--\ref{mamo:end} of Algorithm \ref{alg:local}. 
To determine the personalized submodel, we compute the absolute difference between two consecutive model parameters as
\begin{equation}\label{eq:param-diff}
	\Delta w_n^k = \left| \delta_n^k  \right| 
\end{equation}
where $\delta_n^k :=  w_n^k - w_n^{k+1} \in \mathbb{R}^d$.

Note that the values of $\Delta w_n^k$ reflect the sensitivity of individual model parameters to local data. 
Model parameters exhibiting larger differences, as defined in \eqref{eq:param-diff}, are more strongly influenced by the unique data distribution of the client and therefore are better suited for personalization.
Accordingly, we select the top-$p$ parameters in $\Delta w_n^k$ with the highest magnitudes and designate them as personalized by updating the mask as
\begin{equation}\label{eq:mask-update}
m_n^{k+1}[i] = 
	\begin{cases}
		1, & \text{if } \Delta w_n^k[i] \text{ in top-$p$ }  \\
		m_n^k[i], & \text{otherwise.}
	\end{cases}
\end{equation} 

To prevent over-personalization, we impose a constraint on the size of personalized submodel by introducing a personalization budget $\gamma \in [0,1]$, such that $\norm{m_n^k} \leq \gamma d$ is enforced throughout training with $d$ denoting the dimension of model parameters.
After applying the mask in \eqref{eq:mask-update}, we obtain the updated personalized and shared submodels as 
\begin{subequations}\label{eq:param-partition}
\begin{IEEEeqnarray}{rCl}
p_n^{k+1} & = & w_n^{k+1} \circ m_n^{k+1}, \label{eq:param-partition-p}\\
g_n^{k+1} & = & w_n^{k+1} \circ (\mathbf{1} - m_n^{k+1}). \label{eq:param-partition-g}
\end{IEEEeqnarray}
\end{subequations}

Based on \eqref{eq:param-partition}, each client $n$ uploads the shared submodel $g_n^{k+1}$ to the server for aggregation, while retaining the personalized submodel $p_n^{k+1}$ locally to preserve client-specific information.
The client-specific selection process of PWPM module enables each client $n$ to progressively specialize a submodel that aligns with its local data distribution, while still contributing to global knowledge sharing through the remaining shared submodel.

\subsection{Contribution-Oriented Weighted Module}
While the PWPM and MAMO modules enable clients to locally personalize their models and optimize personalized and shared submodels via different optimization trajectories, the merits of FL ultimately rely on aggregating knowledge across clients. 
However, in heterogeneous networks, the aggregation rule requires accurate weights for all clients when data distributions and training dynamics vary significantly. 
Vanilla FL aggregates local models via heuristic or data-volume-based weighted averaging without considering the actual contribution per client's update.
However, such aggregation rules implicitly assume all updates are equally informative, which are rarely satisfied in PFL, where the clients may capture different statistical distribution of local datasets. 
To incorporate each client’s contribution to  global aggregation, we propose the COWA module that assesses both the informativeness and uniqueness of each client’s update.
More specifically, we quantify contribution of each client from the two complementary subspaces, i.e., gradient discrepancy subspace and prediction contribution subspace.

\paragraph{Gradient Score}
In the PFL setting, a client's model update that deviates from the average direction may capture rare or underrepresented data patterns specific to that client. To quantify the directional novelty of each update, we compute the angular deviation between a client’s local model update and the average update of the remaining clients.
The gradient contribution score is defined as 
\begin{equation}\label{gradient2}
\Gamma_n^{\text{grad}} = 1 - \cos(\delta_n^k,  \bar{\delta}_{-n}^k)
\end{equation}
where leave-one-out average direction $\bar{\delta}_{-n}^k$ is defined as $\bar{\delta}_{-n}^k := \nicefrac{(\delta^k -\alpha_n^{k} \delta_n^{k} )}{(1 - \alpha_n^{k} )}$ with $\delta^k := w^{k-1} - w^k$.

Note that a higher gradient contribution score $\Gamma_n^{\text{grad}}$ indicates that the client $n$ follows a distinct update direction, which indicates that the client has the potential to contribute complementary knowledge to the shared model.

\paragraph{Prediction Score}
The value of a client’s update lies not only in enhancing its own performance, but also in its potential to assist other clients in adapting to their local data distributions.
To capture cross-client information contribution in the data space, we evaluate the performance of the aggregated model $w_{-n}^k$ that is obtained by excluding local training data of client $n$ as 
\begin{equation}\label{space2}
\Gamma_n^{\text{data}} = \mathbb{E}_{\zeta_n \sim \mathcal{M}_n} \left[ \ell(w_{-n}^k; \zeta_n) \right]
\end{equation}
where the leave-one-out average model $w_{-n}^k$ is obtained as $w_{-n}^{k}  =\nicefrac{(w^k-\alpha_{n}^{k} w_{n}^k)}{(1-\alpha_{n}^{k})}$.

When the induced model exhibits poor performance on the data of client $n$ (e.g., higher value of \eqref{space2}), the client $n$ provides complementary rather than redundant information to the global aggregation.
In other words, a lower prediction error \eqref{space2} indicates that the model from client $n$ generalizes well to other clients' data distributions and is beneficial for the other clients.

\paragraph{Aggregation Weight}
To quantify the overall contribution of each client’s update, we integrate the contribution scores derived from the gradient discrepancy and the prediction contribution subspaces. 
More specifically, the overall contribution is defined as $\Gamma_n^k = \Gamma_n^{\text{grad}} + \Gamma_n^{\text{pred}}$, where $\Gamma_n^{\text{grad}}$ and $\Gamma_n^{\text{pred}}$ are, respectively, based on \eqref{gradient2} and \eqref{space2}. 
Based on the overall contribution $[\Gamma_n^k]_{n=1}^N$, the aggregation weight is then obtained by normalizing $\Gamma_n^k$ as 
\begin{equation}\label{eq:weight}
	\alpha_n^k = \frac{\Gamma_n^k}{\sum_{i=1}^N \Gamma_i^k}, n = 1, \ldots, N.
\end{equation}

Note that the aggregation weights $[\alpha_n^k]_{n=1}^N$ are applied exclusively to the aggregation of the shared submodel components $g_n^k$. 
The proposed COWA module ensures that clients exhibiting more distinctive optimization directions or possessing more transferable knowledge exert greater influence in the aggregation process. 
Our design promotes robust personalized and adaptive model improvements under heterogeneous distributions.

\section{Numerical Experiments}

In this section, we demonstrate the effectiveness of the proposed CO-PFL framework through comparison with existing methods and provide comprehensive experimental results. We first describe the experimental setup. Then, we report the results under different heterogeneous data scenarios, datasets, FL configurations, and baselines. Finally, we conduct extensive ablation studies on CO-PFL, including ablation studies on different personalized parameters, client sample sizes, and contribution scores.

\subsection{Experimental Setup}

\begin{table*}[t]
\centering
\setlength{\tabcolsep}{5pt}          
\renewcommand{\arraystretch}{1.3}    
\fontsize{10}{10}\selectfont           
\caption{Personalized accuracy (\%) of different methods on CIFAR10, CIFAR10C, CINIC10, and Mini-ImageNet with varying numbers of clients.}
\label{tab:my-table-all}
\begin{tabular}{l|cc|cc|cc|cc}
\toprule
\multirow{2}{*}{\textbf{Methods}} & 
\multicolumn{2}{c|}{\textbf{CIFAR10}} & 
\multicolumn{2}{c|}{\textbf{CIFAR10C}} & 
\multicolumn{2}{c|}{\textbf{CINIC10}} & 
\multicolumn{2}{c}{\textbf{Mini-ImageNet}} \\ 
\cline{2-9}
 & \textbf{10 Clients} & \textbf{50 Clients} 
 & \textbf{10 Clients} & \textbf{50 Clients} 
 & \textbf{20 Clients} & \textbf{50 Clients} 
 & \textbf{20 Clients} & \textbf{50 Clients} \\ 
\midrule
Local Only  & 74.60 & 67.63 & 66.75 & 71.16 & 65.40 & 66.10 & 33.31 & 33.01 \\
FedAvg      & 62.45 & 70.32 & 68.40 & 14.00 & 66.90 & 59.63 & 31.99 & 34.69 \\
FedAvg+FT   & 69.90 & 74.20 & 70.05 & 66.88 & 69.25 & 66.66 & 33.44 & 35.57 \\
\midrule
LG-FedAvg   & 78.30 & 76.04 & 75.70 & 69.72 & 71.40 & 70.45 & 33.38 & 33.26 \\
FedPer      & 79.80 & 75.26 & 70.10 & 68.87 & 70.98 & 68.27 & 34.53 & 33.60 \\
FedPAC      & 79.25 & 75.51 & 76.40 & 72.61 & 69.05 & 71.70 & 34.29 & \textbf{42.87} \\
FedSelect   & 80.09 & \textbf{78.68} & 76.34 & 75.43 & 72.67 &  73.02& 37.24 & 35.21 \\
\midrule
\textbf{CO-PFL} & \textbf{82.86} & 78.06 & \textbf{79.42} & \textbf{75.54}  & \textbf{73.82} & \textbf{74.45}& \textbf{38.76} & 40.79 \\
\bottomrule
\end{tabular}
\end{table*}

\noindent \textit{1) \textit{Datasets and Models:}} We use four practical datasets, e.g., CIFAR10~\cite{krizhevsky2009learning}, CIFAR10C~\cite{hendrycks2019benchmarking}, CINIC10~\cite{DBLP:journals/corr/abs-2007-13518}, and Mini-ImageNet~\cite{vinyals2016matching} (abbreviated as M-ImageNet), considering both label and feature shift. 
CIFAR10, CIFAR10C, and CINIC10 are all 10-class image classification tasks, while M-ImageNet is a 100-class image classification task.
We follow the experimental protocol in~\cite{Lottery, tamirisa2024fedselect}. In the label shift scenario, a randomly initialized ResNet-18 is trained on CIFAR10 and Mini-ImageNet, where distribution shifts occur primarily at the label level. In contrast, in the label-feature shift scenario, we train the same model on CIFAR10C and CINIC10, which additionally involve distributional shifts in both labels and input features.
\vspace{0.5ex}

\noindent \textit{2) \textit{Heterogeneous Data Setups:}} We examine label/feature
distribution shift in heterogeneous data and refer to them as label/feature shift. To simulate label skew, we divide all dataset into data shards with the same sample number for clients \cite{tamirisa2024fedselect}.
In all experiments, training data are distributed to clients in a heterogeneous manner. 
Specifically, we adopt a label-partitioning strategy in which each client receives data from only $s$ classes, with the number of training samples per class limited by a fixed bound $\mathcal{M}_{\text{bound}}$.
For the main experiments, we set $s=2$ and $\mathcal{M}_{\text{bound}}=50$ on CIFAR10, CIFAR10C, and CINIC10, which means that each client is assigned 50 training samples from each of two classes. 
For M-ImageNet, we employ a more diverse setting with $s=10$ and $\mathcal{M}_{\text{bound}}=50$, allowing each client to access 50 training samples from 10 randomly selected classes.
For evaluation, test data are partitioned in the same class-restricted manner as the training set. Each client is assigned 100 test samples per class, i.e., $\mathcal{M}_{\text{bound}}=100$ for all test sets.

\vspace{0.5ex}

\noindent \textit{3) \textit{FL Setup and Baselines:}}  For CIFAR10 and CIFAR10C, the number of clients is set to $N = |C| = 10$, with each client assigned $s = 2$ classes. 
For M-ImageNet and CINIC10, we configure the number of clients as $N = 20$. Each client is assigned data from $s = 10$ classes in M-ImageNet and $s = 2$ classes in CINIC10. Furthermore, to evaluate the performance of CO-PFL in larger-scale federated environments, we conduct additional comparative experiments with 50 clients, and the corresponding results are summarized in Table\ref{tab:my-table-all}}.
In the main experiments, the local batch size is set to 32 and the number of local training iterations is fixed to 1. 
For CO-PFL, the learning rate $\eta$ is selected from $\{1\times10^{-4}, 1 \times 10^{-5}\}$ across all experiments. 
To ensure fair comparison, we tune the learning rate for all baseline methods from $\{1 \times 10^{-2}, 1 \times 10^{-3}, 1 \times 10^{-4}\}$ and report their best performance.
Unless otherwise specified, we adopt the hyperparameter configurations in FedPAC~\cite{xupersonalized} for all baseline methods.

We compare the proposed CO-PFL with a comprehensive set of baseline approaches that span multiple personalization paradigms in federated learning. 
Local Only serves as the lower bound, where each client trains its model independently using local data, without any communication or collaboration across clients.
FedAvg~\cite{mcmahan2017communication} is the standard algorithm that learns a consensus model by averaging client updates; its personalized variant, FedAvg+FT, performs local fine-tuning on each client after global training.
Moreover, We include parameter decoupling methods that partition the model into shared and personalized submodels. FedPer~\cite{arivazhagan2019federated} keep a shared feature extractor and learn a personalized classifier head on each client. LG-FedAvg~\cite{liang2020think} adopts the opposite strategy, allowing each client to maintain its own feature extractor while sharing the classifier layer globally. 
We further evaluate against recent adaptive personalization approaches. FedPAC~\cite{xupersonalized} aligns local and global features by coordinating client classifiers via a collaborative prediction mechanism. 
FedSelect~\cite{tamirisa2024fedselect} dynamically determines which parameters to personalize for each client based on local gradient statistics, but aggregates client updates using uniform averaging regardless of their quality or informativeness. These benchmarks provide a strong comparative foundation across local, global, and hybrid personalization strategies. Although recent methods support fine-grained personalization, the relative importance of client contributions is often neglected during aggregation.

\begin{figure*}[t]
\centering
\includegraphics[width=2.05\columnwidth]{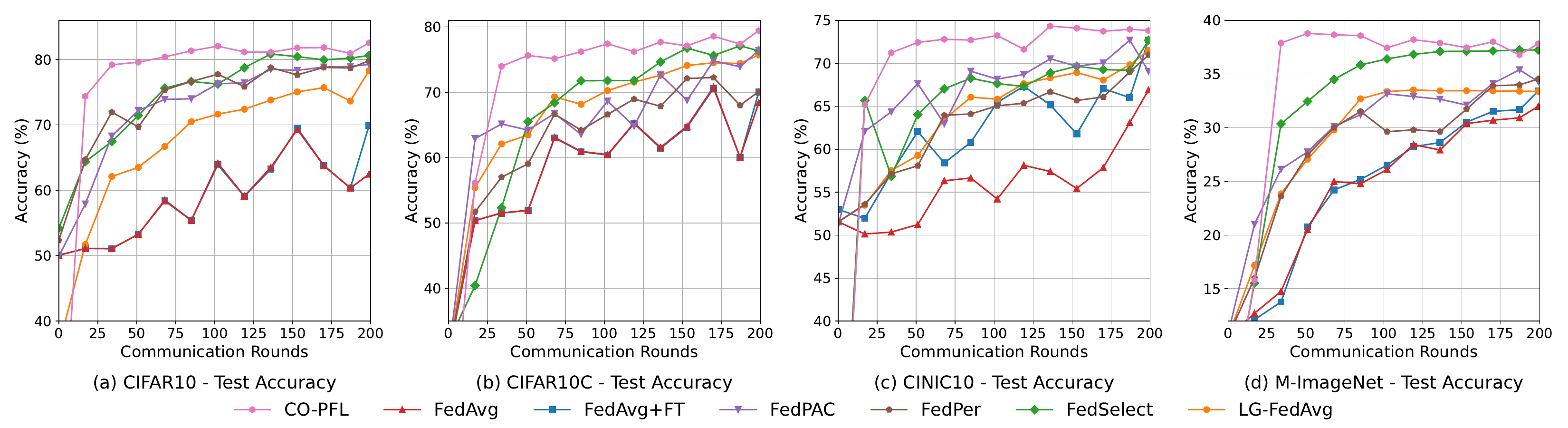} 
\caption{
The test accuracy under few client settings converges with communication rounds in CO-PFL and benchmark.
}    
\label{fig:experiment}
\end{figure*}

\begin{figure*}[t]
\centering
\includegraphics[width=2.05\columnwidth]{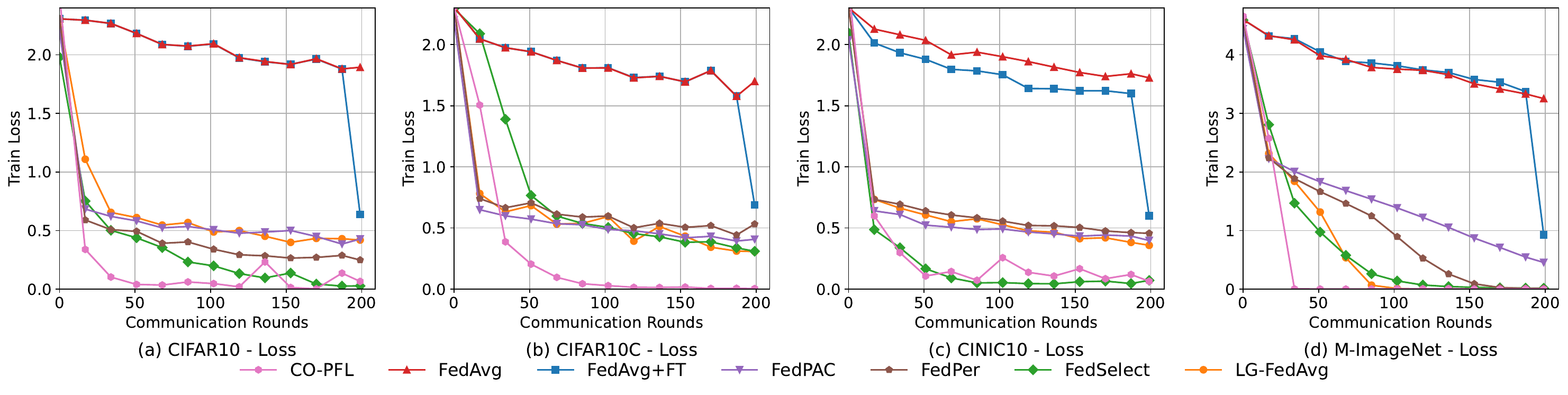} 
\caption{
The train loss under few client settings converges with communication rounds in CO-PFL and benchmark.}    
\label{fig:experiment_loss}
\end{figure*}

\subsection{Numerical Results}

\noindent \textit{1) \textit{Personalization Comparison:}} Table~\ref{tab:my-table-all} presents the personalized accuracy of CO-PFL and competing methods across four datasets under different numbers of clients. Under small scale federal environment, CO-PFL yields a 3.29\% relative improvement in accuracy compared to the benchmarks. Compared to methods such as FedAvg and FedAvg+FT, CO-PFL consistently delivers higher personalized accuracy and more stable convergence. On CIFAR10 and CIFAR10C, CO-PFL achieves a significant performance improvement with a 3.75\% relative improvement in accuracy compared to the benchmarks. On more challenging datasets with greater domain shift, such as CINIC10 and M-ImageNet, CO-PFL is still better than the optimal benchmark with a 2.83\% relative improvement.

When the number of clients increases from 10 (or 20) to 50, the overall performance of most baselines notably declines, reflecting the increasing difficulty of maintaining personalization under more fragmented and heterogeneous data distributions.
In contrast, CO-PFL exhibits remarkable stability, maintaining competitive accuracy across all datasets.
Specifically, on CIFAR10 and CIFAR10C, CO-PFL achieves 78.06\% and 75.54\% accuracy with 50 clients.
On the more challenging CINIC10 and Mini-ImageNet datasets, where domain shift and client diversity are more pronounced, CO-PFL still achieves the best performance, with accuracies of 74.45\% and 40.79\%, respectively.
These results highlight the scalability and robustness of CO-PFL in handling increasingly heterogeneous and large-scale federated environments. Overall, CO-PFL consistently achieves the competitive accuracy across all settings, demonstrating superior adaptability to both small and large scale federated environments.

Figure~\ref{fig:experiment} and Figure~\ref{fig:experiment_loss} respectively show the convergence behavior of CO-PFL and representative baselines in terms of test accuracy and training loss under the small scale federated environment.
Across all four datasets, CO-PFL consistently achieves the fastest and most stable convergence, demonstrating both higher final accuracy and lower training loss than competing methods. Unlike FedAvg, FedPer, and FedPAC, which often exhibit oscillations or early performance saturation under heterogeneous data, CO-PFL converges smoothly within approximately 75 communication rounds.
This improvement can be attributed to the joint effect of its three core modules: the COWA module adaptively weights client updates to mitigate aggregation bias, the PWPM module enables fine-grained parameter personalization, and the MAMO module stabilizes optimization through momentum decoupling. Together, these components facilitate balanced global coordination and local adaptation, enhancing both the robustness and generalization of the overall learning process.

\vspace{0.5ex}

\begin{figure}[t]
\centering
\includegraphics[width=0.9\columnwidth]{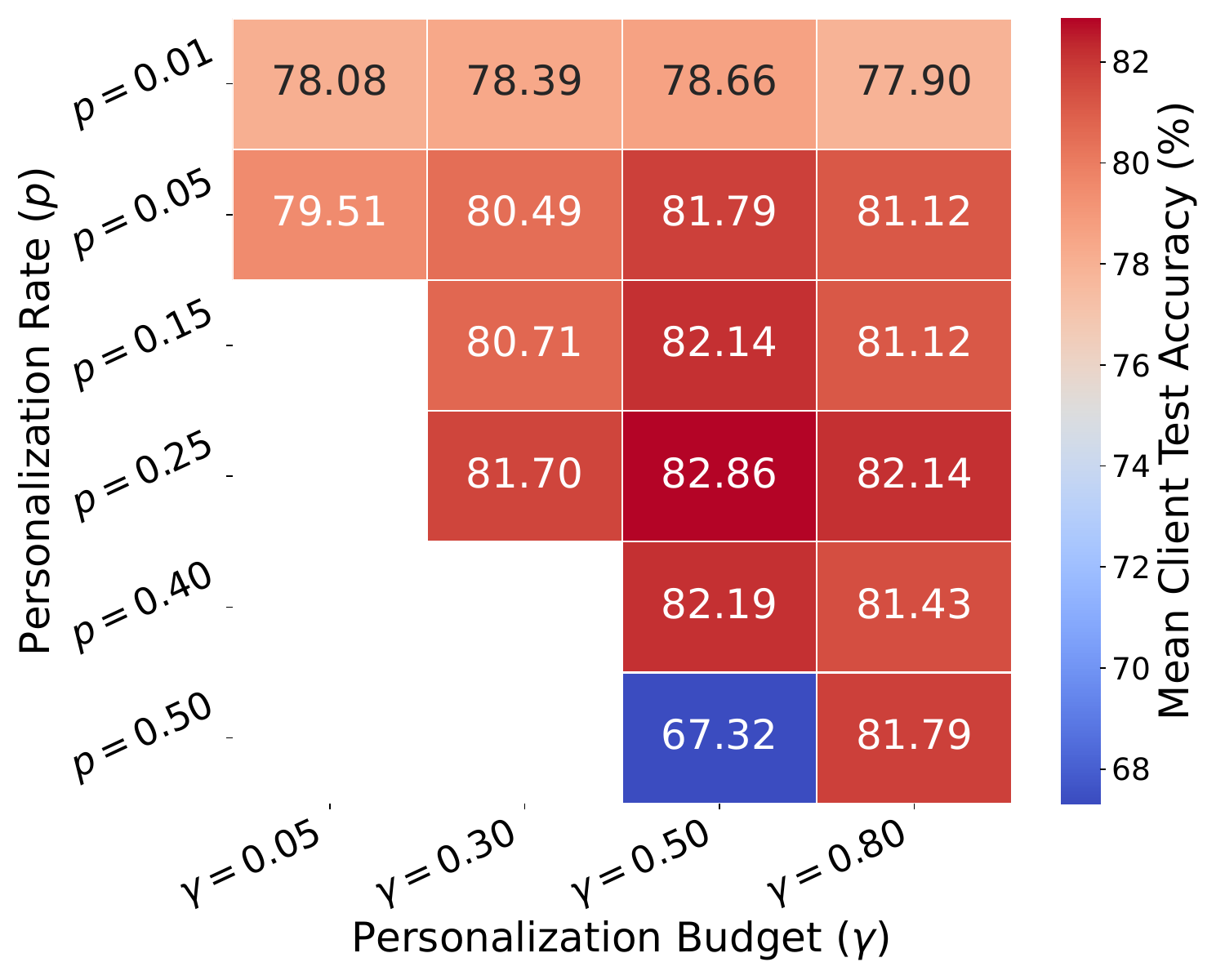} 
\caption{Performance of CO-PFL on CIFAR10 with varying personalization rates $p$ and budgets $\gamma$.}    
\label{fig:heatmap}
\includegraphics[width=0.9\columnwidth]{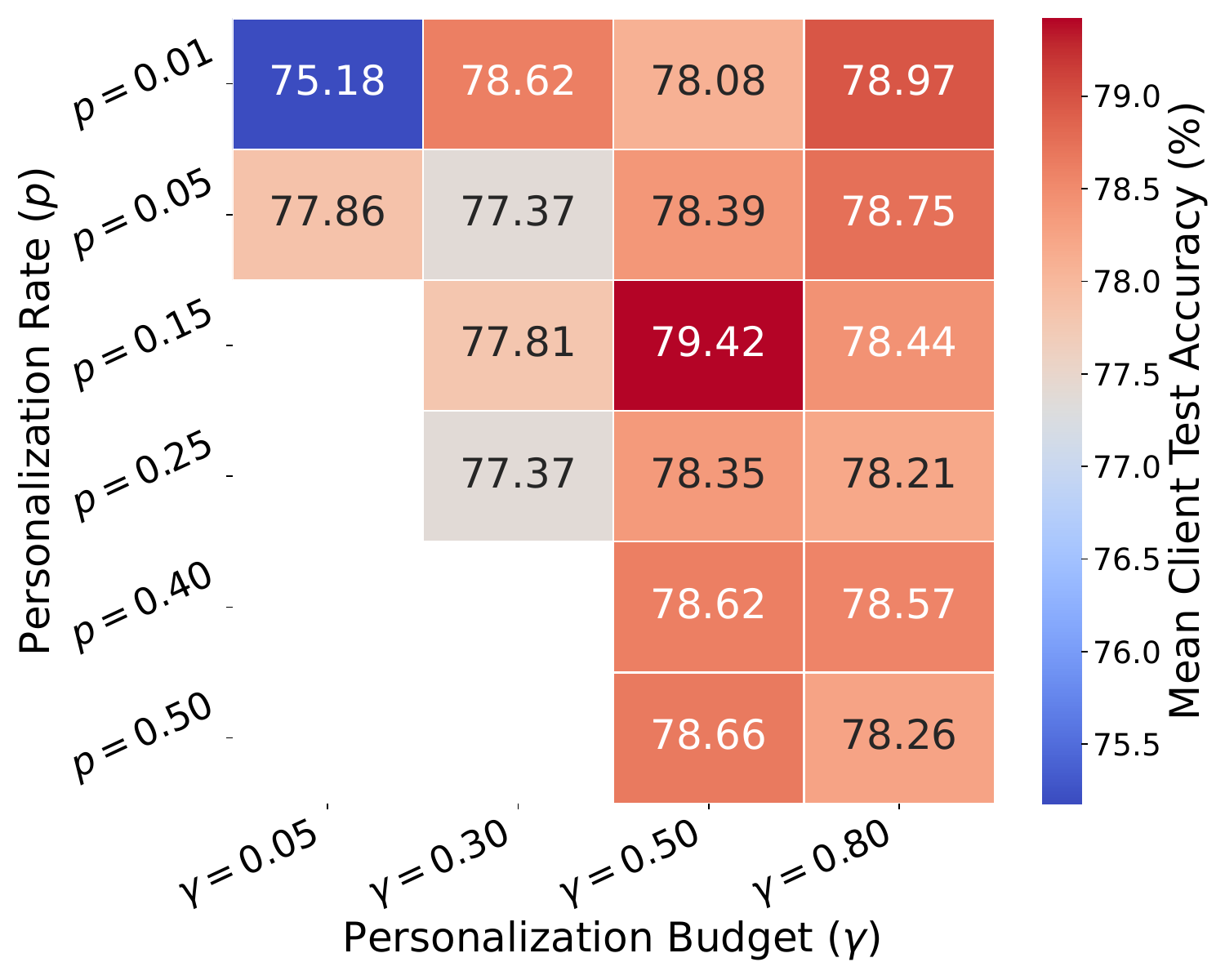} 
\caption{Performance of CO-PFL on CIFAR10C with varying personalization rates $p$ and budgets $\gamma$.}    
\label{fig:heatmap_C}
\end{figure}

\noindent \textit{2) \textit{Performance on different personalized parameters:}} 
Figure~\ref{fig:heatmap} and Figure~\ref{fig:heatmap_C} highlights the impacts of personalization rate $p$ and personalization budget $\gamma$ on the convergence accuracy under small acale FL environment, where $p \in \{0.01, 0.05, 0.15, 0.25, 0.40, 0.50\}$ and $\gamma \in \{0.05, 0.30, 0.50, 0.80\}$ to examine the sensitivity of the model to different personalization configurations.

We first analyze the sensitivity of CO-PFL to personalization rate $p$ and budget $\gamma$ on the CIFAR10 dataset. As summarized in Figure \ref{fig:heatmap}, extreme configurations lead to degraded performance. For instance, when $p=0.01$ and $\gamma=0.05$, CO-PFL essentially degenerates to vanilla FedAvg, reaching only 78.08\%. Conversely, with $p=0.50$ and $\gamma=0.50$, the model behaves almost like fully local training and suffers from overfitting and instability, yielding only 67.32\%. In contrast, a moderate configuration with $p=0.25$ and $\gamma=0.50$ achieves the best accuracy of 82.86\%, indicating that balanced personalization effectively trades off global coordination and local adaptation.

Figure~\ref{fig:heatmap} visualizes the test accuracy across different combinations of $p$ and $\gamma$. The heatmap corroborates the findings: accuracy peaks at moderate values, whereas extremely small $p$ values provide little personalization gain and overly large $p$ values cause instability. While increasing $\gamma$ expands the scope of personalization, its benefit diminishes without a controlled $p$. These results suggest that CO-PFL performs best when a balanced personalization rate and budget are jointly applied to maintain robustness and generalization.

We further extend analysis to the CIFAR10C dataset, where distributional shifts are more severe. As reported in Figure~\ref{fig:heatmap_C}, the best accuracy of 79.42\% is achieved at $p=0.15$ and $\gamma=0.50$, again demonstrating that moderate personalization offers the most effective trade-off between global knowledge sharing and local adaptation. By contrast, excessive personalization (e.g., $p=0.50$) degrades accuracy to 78.66\%, while insufficient personalization ($p=0.01, \gamma=0.05$) further drops performance to 75.18\%.

Overall, across both datasets, CO-PFL demonstrates consistent trends: extreme parameter choices either collapse to near FedAvg performance or cause overfitting, whereas moderate values of $p$ and $\gamma$ enable CO-PFL to maintain robustness and achieve superior accuracy under heterogeneous distributions.

\vspace{0.5ex}

\begin{figure}[t]
\centering
\includegraphics[width=0.95\columnwidth]{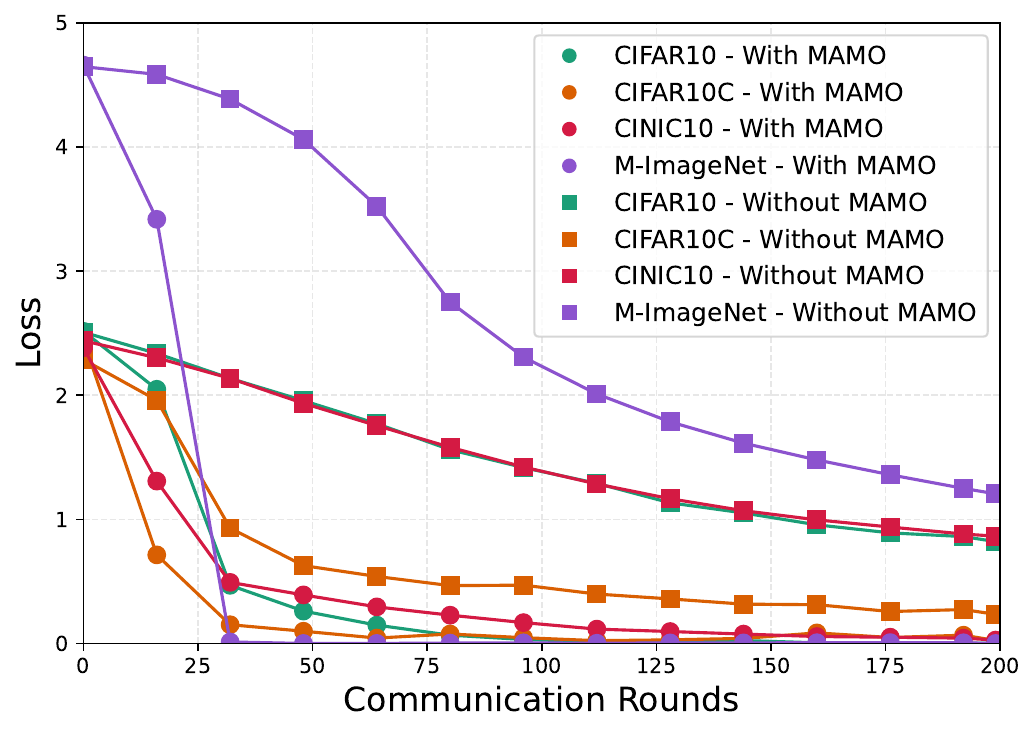} 
\caption{The convergence behaviors for CO-PFL with and without MAMO module.}    
\label{fig:experimentmamo}
\end{figure}

\noindent \textit{3) Performance of the MAMO Module:}
Figure~\ref{fig:experimentmamo} highlights the importance of tailoring the optimization trajectory to structurally decoupled submodels in PFL. By restricting momentum updates to the selected submodel, the proposed MAMO module effectively alleviates gradient interference between shared and personalized parameters. As illustrated in the figure, models equipped with MAMO consistently converge faster and more stably across all datasets. In particular, on CIFAR10 and CIFAR10C, the training loss with MAMO drops sharply within the first 50–75 communication rounds and quickly stabilizes near zero, whereas the versions without MAMO converge more slowly and remain at noticeably higher loss values throughout training. A similar trend is observed on CINIC10 and Mini-ImageNet, where MAMO substantially reduces oscillations and improves convergence smoothness under more heterogeneous and challenging data distributions.

These consistent improvements demonstrate that MAMO not only accelerates convergence but also enhances stability in scenarios with severe heterogeneity and limited client data. The results confirm that isolating the momentum of shared and personalized submodels prevents harmful gradient entanglement, thereby enabling more reliable personalization. Overall, the integration of MAMO into CO-PFL significantly improves both robustness and communication efficiency of federated training.

\subsection{Ablation studies of the COWA}

In this ablation study, we examine the effectiveness of the two components in the contribution score calculation module, namely gradient-based contribution and prediction-based contribution. The results across four datasets are summarized in Table~\ref{tab:ablation}. For all experiments, we fixed the learning rate to $1\times10^{-4}$ and adopted the optimal personalization parameters of $p=\{0.15,0.25\}$ and $\gamma=0.50$.

The results consistently demonstrate that the joint use of both components yields the best performance. On CIFAR10, incorporating both contributions achieves the highest accuracy of 82.86\%. Removing either element leads to performance degradation: 81.43\% with gradient alone and 82.23\% with prediction alone. This indicates that prediction signals provide a slightly stronger standalone effect, while the gradient component adds complementary benefits. A similar pattern appears on CIFAR10C, where the combined design reaches 79.42\%, outperforming the gradient-only (79.20\%) and prediction-only (78.97\%) settings.

The advantage of combining both components is even more pronounced on more challenging datasets. For Mini-ImageNet, the complete module improves accuracy to 38.76\%, compared with 37.01\% (gradient only) and 37.88\% (prediction only). On CINIC, which involves stronger distributional noise, the combination further raises accuracy to 73.82\%, significantly above the single-component variants (72.66\% and 72.84\%).

Taken together, these results highlight the complementary nature of gradient-based and prediction-based contributions. While each offers benefits individually, their integration ensures more reliable estimation of client contribution, thereby enhancing model robustness and generalization across diverse and heterogeneous datasets.

\subsection{Exploratory Study}

\begin{table}[t]
\centering
\setlength{\tabcolsep}{4pt}         
\renewcommand{\arraystretch}{1.2}   
\fontsize{9}{10}\selectfont       
\caption{Ablation experiment on the contribution score calculation module.}
\renewcommand{\arraystretch}{1.15}  
\begin{tabular}{c|c|c|c}
\hline
\multirow{2}{*}{\textbf{Dataset}} & \multicolumn{2}{c|}{\textbf{Components}} & \multirow{2}{*}{\textbf{Accuracy(\%)}} \\ \cline{2-3} 
 & \textbf{Gradient} & \textbf{Prediction} & \\ \hline
\multirow{4}{*}{\textbf{CIFAR10}} &  \ding{55} & \ding{55} & 80.62 \\ 
 & \ding{55} & \checkmark & 82.23 \\
 & \checkmark &  \ding{55} & 81.43 \\
 & \checkmark & \checkmark & \textbf{82.86} \\ \hline 
\multirow{4}{*}{\textbf{CIFAR10C}} & \ding{55} & \ding{55} & 78.71 \\ 
 & \ding{55} & \checkmark & 78.97 \\
 & \checkmark & \ding{55} & 79.20 \\
 & \checkmark & \checkmark & \textbf{79.42} \\ \hline  

\multirow{4}{*}{\textbf{M-ImageNet}} & \ding{55} & \ding{55} & 36.77 \\ 
 & \ding{55} & \checkmark & 37.88 \\
 & \checkmark & \ding{55} & 37.01 \\
 & \checkmark & \checkmark & \textbf{38.76} \\ \hline 

 \multirow{4}{*}{\textbf{CINIC10}} & \ding{55} & \ding{55} & 46.63\\ 
 & \ding{55} & \checkmark & 72.84 \\
 & \checkmark & \ding{55} & 72.66 \\
 & \checkmark & \checkmark & \textbf{73.82} \\ \hline

\end{tabular}
\label{tab:ablation}
\end{table}

\begin{table}[t]
\centering

\caption{Comparison of performance (\%) on CIFAR10 when varying the training data size $\mathcal M_{bound} $ per client. Each client was assigned with 2 classes.}
\renewcommand{\arraystretch}{1.2}  
\begin{tabular}{lcccc}
\hline
Methods      & $\mathcal M = 10$    & $\mathcal M = 50$    & $\mathcal M = 100$   & $\mathcal M = 200$   \\ \hline
{ Local Only} & {  50.96} & {  74.60}  & {  75.15} & {  81.70} \\
  
{  FedAvg}     & {  24.30}  & {  62.45} & {  76.85} & {  82.30}  \\
{  FedAvg+FT}  & {  53.70}  & {  69.90}  & {  79.20}  & {  83.05} \\ \hline
  
{  LG-FedAvg}  & {  64.90}  & {  78.30}  & {  82.70}  & {  85.20}  \\
{  FedPer}     & {  61.35} & {  79.80}  & {  79.90}  & {  84.65} \\
  
{  FedPAC}     & {  65.10}  & {  79.25} & {  80.45} & {  84.30}  \\
{  FedSelect}  & {  70.31} & {  80.09} & {  82.77} & {  85.13} \\
  
{  \textbf{CO-PFL}}     & {  \textbf{72.28}} & {  \textbf{82.86}} & {  \textbf{83.44}} & {  \textbf{86.07}} \\ \hline
\end{tabular}

\label{tab:my-table21}
\end{table}

\begin{table}[t]
\renewcommand{\arraystretch}{1.2} 
\caption{Comparison of performance (\%) on CIFAR10C when varying the training data size $\mathcal M_{bound} $ per client. Each client was assigned with 2 classes.}
\begin{tabular}{lcccc}
\hline
Methods                           & $\mathcal M = 10$    & $\mathcal M = 50$   & $\mathcal M = 100$   & $\mathcal M = 200$   \\ \hline
{Local Only} & {64.05} & {66.75} & {79.30}  & {80.30}  \\

{FedAvg}     & {52.55} & {68.40} & {73.45} & {76.70} \\
{FedAvg+FT}  & {55.70}  & {70.05} & {75.15} & {79.95} \\ \hline

{LG-FedAvg}  & {63.05} & {75.70}  & {79.15} & {84.85} \\
{FedPer}     & {67.00}    & {70.10}  & {80.50} & {85.20}  \\

{FedPAC}     & {64.45} & {76.40}  & {77.26} & {83.00} \\
{FedSelect}  & {\textbf{70.49}}      & {76.34} & {78.30}      & {84.15}      \\

{\textbf{CO-PFL}} &
  {70.27} &
  {\textbf{79.42}} &
  {\textbf{83.04}} &
  {\textbf{85.76}} \\ \hline
\end{tabular}
\centering

\label{tab:my-table22}
\end{table}

We evaluate the performance of CO-PFL and benchmarks on CIFAR10 and CIFAR10C under varying amounts of training data per client in small scale FL environment, with each client assigned 2 classes to maintain consistent label space heterogeneity. 
The range of sample sizes in the training set is $\mathcal{M}_{bound} \in \{20,100,200,400\}$, while the size of the test set remains 200 and there is no overlap with the training set. 

As shown in Tables~\ref{tab:my-table21} and~\ref{tab:my-table22}, the performance of all benchmarks improves with increased local data. However, CO-PFL consistently achieves the best results across all regimes, particularly under scarce and highly heterogeneous conditions. In the scarce regime ($\mathcal{M}_{bound} = 10$), CO-PFL outperforms all benchmarks by a notable margin. On CIFAR10, it achieves 72.28\%, exceeding the second-best method (FedSelect, 70.31\%) by 1.97\%; on CIFAR10C, it reaches 70.27\%, nearly matching the best baseline (FedSelect, 70.49\%) despite the added feature distribution shifts. Experimental results demonstrate the advantage of contribution-oriented aggregation and selective personalization in mitigating optimization bias caused by scarce and heterogeneous data.

At $\mathcal{M}_{bound} = 100$, CO-PFL achieves 83.44\% and 83.04\% on CIFAR10 and CIFAR10C respectively, outperforming the FedSelect by 0.67\% and 4.74\%. Notably, CO-PFL continues to scale well with larger datasets, reaching 86.07\% on CIFAR10 and 85.76\% on CIFAR10C at $\mathcal{M}_{bound} = 200$, reflecting strong generalization even under less constrained local training conditions.
The results in CIFAR10C further highlight CO-PFL's robustness under distributional shift, where both label and feature shifts coexist. Although most benchmarks exhibit reduced or unstable performance in this setting, CO-PFL maintains consistent improvements across all data scales, showcasing its ability to unify personalization and collaboration across heterogeneous clients.

Experimental results demonstrate that integrating the COWA, PWPM, and MAMO modules enables CO-PFL to adapt effectively across diverse heterogeneous federated settings, enhancing optimization robustness and generalization in large scale FL environments.

\section{Conclusions}

In this work, we introduced CO-PFL, a comprehensive framework that effectively bridges the critical gap between local model personalization and global collaborative learning in federated environments. By moving beyond naive averaging, our proposed COWA module leverages a dual assessment of gradient direction and prediction deviation to dynamically quantify the true utility of each client's update, enabling a more informed and bias-resistant aggregation. This contribution-aware core is powerfully supported by the PWPM, which allows for fine-grained, dynamic parameter selection tailored to local data characteristics, and the Mask-Aware Momentum Optimization, which ensures stable convergence by decoupling the optimization of shared and personalized parameters. Together, these two modules of COWA and PWPM form a unified and practical solution that directly addresses the challenges of statistical heterogeneity and scarce data. Extensive experiments on benchmark datasets confirm that CO-PFL not only mitigates aggregation bias and enhances global coordination, but also consistently achieves state-of-the-art personalization performance, establishing a robust new paradigm for practical federated learning systems.

\bibliographystyle{IEEEtran}   
\bibliography{ref}          

\begin{IEEEbiographynophoto}{Ke Xing}
received the B.S. degree in computer science from Central South University, Changsha, China, in 2020, and the M.S. degree in computer science from Lanzhou University, Lanzhou, China, in 2024. He is currently pursuing the Ph.D. degree in computer science and technology at Beijing Institute of Technology, Beijing, China. His research interests include federated learning, differential privacy, and large-model optimization.

\end{IEEEbiographynophoto}

\begin{IEEEbiographynophoto}{Yanjie Dong (Member, IEEE)}
   is currently an Associate Professor and Assistant Dean of Artificial Intelligence Research Institute, Shenzhen MSU-BIT university. Dr. Dong obtained his PhD and MASc degree from The University of British Columbia, Canada, in 2020 and 2016, respectively. His research interests focus on the protocol design of energy-efficient communications, machine learning based resource allocation algorithms, and quatum computing technologies. He regularly serves as a member of Technical Program Committee in flagship conferences in IEEE ComSoc.
\end{IEEEbiographynophoto}

\begin{IEEEbiographynophoto}{Xiaoyi Fan (Member, IEEE)}
     is currently a Professor at Shenzhen MSU-BIT University, Shenzhen, China. He received the B.Eng. degree from the Beijing University of Posts and Telecommunications in 2013, and the PhD degree from Simon Fraser University in 2018. From 2018 to 2021, he worked as an honorary postdoctoral research fellow with the Department of Electrical and Computer Engineering, University of British Columbia. He is a recipient of NSERC Postdoctoral Fellowship (2019). He is a recipient of the IEEE ICDCS Distinguished Paper Award (2024) and the IWQoS Best Poster Award (2024). His research interests include the Internet of Things, edge computing, and smart grid.
\end{IEEEbiographynophoto}

\begin{IEEEbiographynophoto}{Runhao Zeng (Member, IEEE)}
    received the PhD degree in software engineering from South China University of Technology, in 2021. He is currently an associate professor at the Artificial Intelligence Research Institute, Shenzhen MSU-BIT University. He has authored or coauthored several peer-reviewed papers on computer vision, machine learning on top-tier conferences and journals, including the Proceedings of NeurIPS, CVPR, ICCV, and TPAMI. His current research interests include machine learning, computer vision, with particular focus on video analysis.

\end{IEEEbiographynophoto}

\begin{IEEEbiographynophoto}{Victor C. M. Leung (Life Fellow, IEEE)}
	is a Distinguished Professor and Dean of Artificial Intelligence Research Institute, Shenzhen MSU-BIT University, China. He is also an Emeritus Professor of Electrical and Computer Engineering and Director of the Laboratory for Wireless Networks and Mobile Systems at the University of British Columbia (UBC), Canada. His research is in the broad areas of wireless networks and mobile systems, and he has published widely in these areas. Dr. Leung is serving as a Senior Editor of the IEEE Transactions on Green Communications and Networking. He is also serving on the editorial boards of the IEEE Transactions on
Cloud Computing, IEEE Transactions on Computational Social Systems, IEEE Access, IEEE Network, and several other journals. He received the 1977 APEBC Gold Medal, 1977-1981 NSERC Postgraduate Scholarships, IEEE Vancouver Section Centennial Award, 2011 UBC Killam Research Prize, 2017 Canadian Award for Telecommunications Research, 2018 IEEE TCGCC Distinguished Technical Achievement Recognition Award, and 2018 ACM MSWiM Reginald Fessenden Award. He co-authored papers that won the 2017 IEEE ComSoc Fred W. Ellersick Prize, 2017 IEEE Systems Journal Best Paper Award, 2018 IEEE CSIM Best Journal Paper Award, and 2019 IEEE TCGCC Best Journal Paper Award. He is a Life Fellow of IEEE, and a Fellow of the Royal Society of Canada (Academy of Science), Canadian Academy of Engineering, and Engineering Institute of Canada. He is named in the current Clarivate Analytics list of “Highly Cited Researchers”.
\end{IEEEbiographynophoto}

\begin{IEEEbiographynophoto}{Xiping Hu (Member, IEEE)}
is currently a Professor with Shenzhen MSU-BIT University, and is also with Beijing Institute of Technology, China. Dr.
Hu received the Ph.D. degree from the University of British Columbia, Vancouver, BC, Canada. Dr.Hu is the co-founder and chief scientist of Erudite Education Group Limited, Hong Kong, a leading language learning mobile application company with over 100 million users, and listed as top 2 language education platform globally. His research interests include affective computing, mobile cyberphysical systems, crowdsensing, social networks, and cloud computing. He has published more than 150 papers in the prestigious conferences and journals, such as IJCAI, AAAI, ACM MobiCom, WWW, and IEEE TPAMI/TMM/TVT/IoTJ/COMMAG.
\end{IEEEbiographynophoto}

\begin{IEEEbiographynophoto}{Dr. M. Jamal Deen (Life Fellow, IEEE)}
is a Distinguished University Professor, Dr. Haykin Distinguished Engineering Professor and Director of the Micro- and Nano-Systems Laboratory (MNSL), McMaster University. As an educator, he won the SM Sze Education Award from IEEE Electron Devices Society (inaugural winner), the Ham Education Medal from IEEE Canada, the McMaster University President’s Award for Excellence in Graduate Supervision, and MSU Macademics’ Lifetime Achievement Award for his exceptional dedication to teaching and significant contribution to student life, the community at large, and academia.
Dr. Deen served as the elected President of the Academy of Science, The Royal Society of Canada in 2015-2017. Currently, he is serving as the inaugural elected Vice President (North) of The World Academy of Sciences, representing the developed countries. His current research interests are nanoelectronics, optoelectronics, nanotechnology, data analytics and their emerging applications to  health and environmental sciences. Dr. Deen’s research record includes more than 930 peer-reviewed articles (about 20\% are invited), two textbooks on “Silicon Photonics- Fundamentals and Devices” and “Fiber Optic Communications: Fundamentals and Applications”, 13 awarded patents of which 7 are/were extensively used in industry, and twenty-seven best paper/poster/presentation awards.
As an undergraduate student at the University of Guyana, Dr. Deen was the top ranked mathematics and physics student and the second ranked student at the university, winning the Irving Adler prize and the Chancellor’s gold medal, respectively. As a graduate student, he was a Fulbright-Laspau Scholar and an American Vacuum Society Scholar. He is a Distinguished Lecturer of the IEEE Electron Device Society for more than two decades now. His awards and honors include the Callinan Award as well as the Electronics and Photonics Award from the Electrochemical Society; a Humboldt Research Award from the Alexander von Humboldt Foundation; the Eadie Medal from the Royal Society of Canada; McNaughton Gold Medal, the Fessenden Medal and the Gotlieb Computer Medal, all from IEEE Canada. In addition, he was awarded the five honorary doctorate degrees in recognition of his exceptional research, scholarly and educational accomplishments, exemplary professionalism and valued services.
Dr. Deen has been elected by his peers as Fellow/Academician of fourteen national academies and professional societies including The Royal Society of Canada; Academician (Foreign Member) of The Chinese Academy of Sciences; The World Academy of Sciences, National Academy of Sciences India; The African Academy of Sciences; The American Physical Society; and The Electrochemical Society (FECS). He was also elected to the Order of Canada, the highest civilian honor awarded by the Government of Canada.
\end{IEEEbiographynophoto}

\vfill

\end{document}